\documentclass[letterpaper, 10 pt, conference]{ieeeconf}  

\IEEEoverridecommandlockouts                              

\usepackage{graphics} 
\usepackage{epsfig} 
\usepackage{times} 
\usepackage{amsmath} 
\usepackage{amssymb}  
\usepackage{makecell}
\usepackage{booktabs}
\usepackage{multirow}
\usepackage{multicol}
\usepackage{here}
\usepackage{kotex}
\usepackage{soul, color}
\usepackage{array}
\usepackage{threeparttable}
\usepackage{hyperref}
\usepackage{stfloats}
\usepackage[table]{xcolor}
\newcommand{\plh}{%
  {\ooalign{$\phantom{0}$\cr\hidewidth$\scriptstyle\times$\cr}}%
}

\title{\LARGE \bf
TriLift: Interpolation-Free Tri-Plane Lifting for Efficient 3D Perception on Embedded Systems
}

\author{
Sibaek Lee, Jiung Yeon and Hyeonwoo Yu
 \thanks{Sibaek Lee, Jiung Yeon and Hyeonwoo Yu are with the Department of Intelligent Robotics, Sungkyunkwan University, Suwon, South Korea. {\tt\small \{lmjlss, wcr12st, hwyu\}@skku.edu}}
 \thanks{
 The code is available at: \href{https://github.com/Lab-of-AI-and-Robotics/TriLift}{https://github.com/Lab-of-AI-and-Robotics/TriLift}}
}

\begin{document}

\maketitle
\thispagestyle{empty}
\pagestyle{empty}

\begin{abstract}
Dense 3D convolutions provide high accuracy for perception but are too computationally expensive for real-time robotic systems.
Existing tri-plane methods rely on 2D image features with interpolation, point-wise queries, and implicit MLPs, which makes them computationally heavy and unsuitable for embedded 3D inference. As an alternative, we propose TriLift, a novel interpolation-free tri-plane lifting and volumetric fusion framework that directly projects 3D voxels into plane features and reconstructs a feature volume through broadcast and summation. This shifts nonlinearity to 2D convolutions, reducing complexity while remaining fully parallelizable. To mitigate spatial information loss inherent in projections, we incorporate a lightweight adaptive positional encoding module that helps bridge the spatial information gap, dynamically recovering fine geometric details with negligible overhead. To capture global context, we add a low-resolution volumetric branch fused with the lifted features through a lightweight integration layer, yielding a design that is both efficient and end-to-end GPU-accelerated. 
To validate the effectiveness of the proposed method, we conduct experiments on classification, completion, segmentation, and detection, and we map the trade-off between efficiency and accuracy across tasks. Results show that classification and completion retain or improve accuracy, while segmentation and detection show a trade-off, significantly reducing computational demand with only a slight decrease in accuracy. On-device benchmarks on an NVIDIA Jetson Orin Nano confirm robust real-time throughput, demonstrating the suitability of the approach for embedded robotic perception.
\end{abstract}



\section{INTRODUCTION}
\begin{figure}[t]
    \centering
    \includegraphics[width=0.48\textwidth]{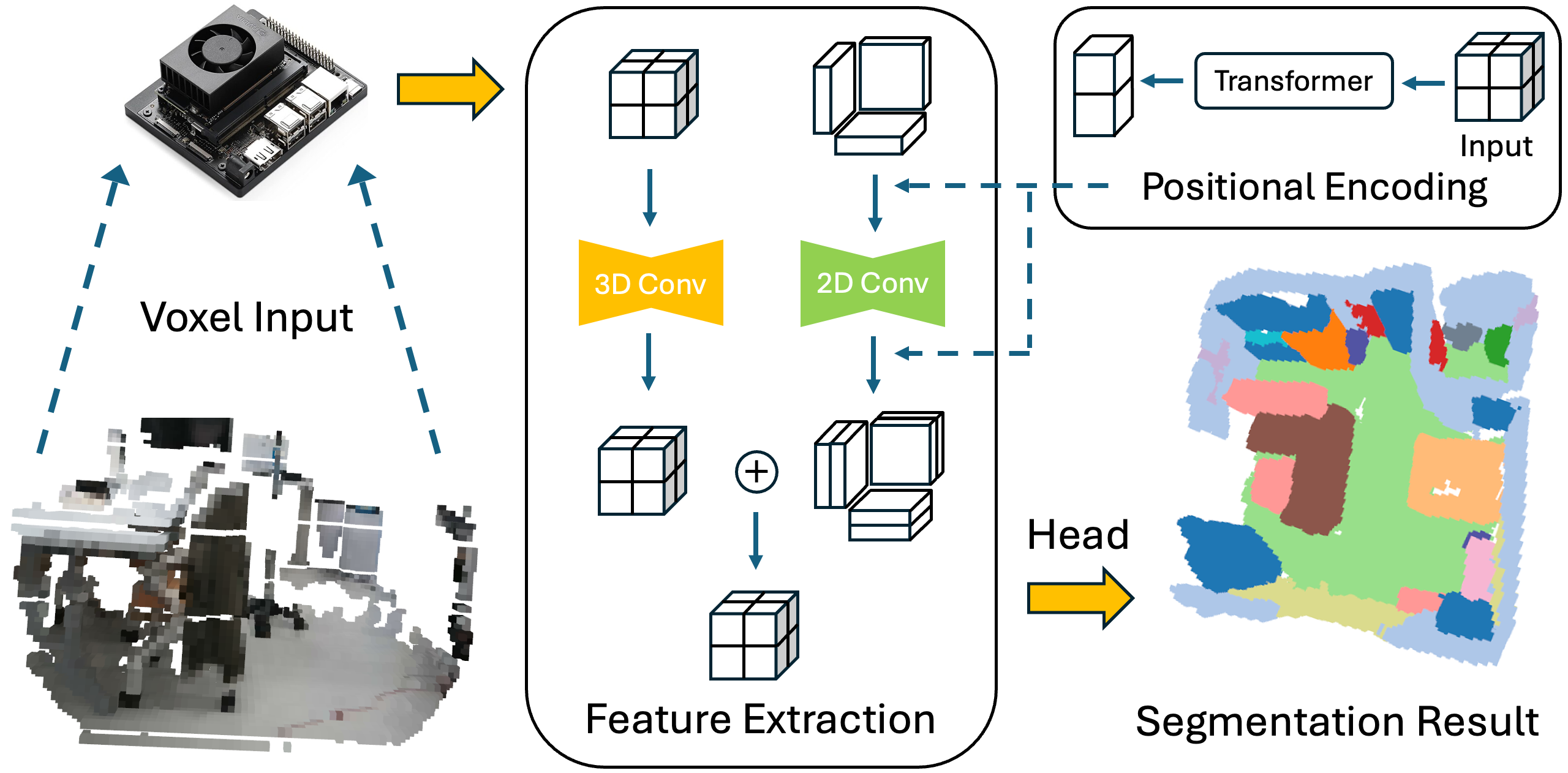}
    \caption{Hardware-efficient 3D perception framework for embedded systems. Our approach enables real-time 3D inference by employing an interpolation-free tri-plane lifting scheme. The architecture minimizes memory-access overhead by fusing a high-speed 2D stream with a coarse 3D branch, while adaptive positional encoding ensures high geometric fidelity with negligible computational cost.}
    \label{fig:intro_setup} 
\end{figure}

Real-time, precise 3D perception is a critical foundation for autonomous robots to understand and interact with their surroundings. We address this with an efficient framework, illustrated in Figure~\ref{fig:intro_setup}. However, as voxel resolution increases, this cubic growth in computational complexity and memory usage quickly exceeds the limits of embedded systems, making real-time processing impossible~\cite{tatarchenko2017octree, liu2019point}.
To address these challenges, methods typically exploit spatial sparsity or project 3D volumes onto 2D planes. Sparse convolution techniques~\cite{graham2017submanifold, graham20183d} selectively perform computation only on non-empty voxel locations, thereby avoiding wasted operations in unoccupied space.
Approaches such as MinkowskiNet~\cite{choy20194d} focus processing on non-empty regions, but they remain computationally expensive due to their reliance on 3D convolutions. While this sparsity accelerates binary occupancy inputs, its advantage vanishes on continuous-density fields, where most voxels carry nonzero values.
Projection-based methods~\cite{chan2022efficient, shue20233d, lee2024semcity, huang2023tri} construct an axis-aligned tri-plane representation from 2D image features or latent volumes, and then interpolate features from the three planes to query arbitrary 3D points. These features are typically processed by implicit MLPs for downstream 3D tasks, such as generative scene synthesis or occupancy prediction. While this scheme leverages the efficiency of 2D convolutions, it still depends on point-wise queries and interpolation, leading to high computational overhead and making it less suitable for real-time embedded reasoning.

In this work, we revisit the tri-plane representation as a lightweight backbone for efficient 3D voxel reasoning. Unlike prior approaches that applied tri-planes for generative modeling or occupancy prediction, our design targets real-time inference on embedded robotic systems. We propose TriLift, an interpolation-free lifting scheme that derives feature tri-planes from voxel projections with 2D convolutions, and restores a 3D feature volume via broadcast and summation. This design removes the need for interpolation, point-wise queries, and implicit MLPs. We further strengthen this backbone by adding a low-resolution volumetric branch and integrating it with the lifted features through a lightweight channel-mixing step. This hybrid design preserves coarse global context while maintaining an end-to-end GPU accelerated structure, enabling both efficiency and accuracy gains over interpolation-based tri-plane methods. In addition, we include an adaptive positional encoding as an auxiliary modulation. Unlike fixed sinusoidal encodings~\cite{liu2018intriguing} or voxel-wise MLPs~\cite{liu2022petr}, our module generates adaptive embeddings that capture inter-slice relationships. 
Specifically, our transformer-based module summarizes axis-wise structural context into 1D tokens and applies them through a two-stage additive modulation, both before projection and after reconstruction. This allows the framework to mitigate the spatial information loss inherent in 2D projections by learning data-dependent geometric cues.
Because its computational overhead is negligible (less than 0.1\% of the backbone's FLOPs), it provides an economical way to enhance reconstruction fidelity by recovering fine geometric detail.

We evaluate our method on a diverse set of 3D vision tasks, including shape classification and completion on ModelNet40~\cite{wu20153d}, semantic segmentation on ScanNet~\cite{dai2017scannet} and Stanford3D~\cite{armeni20163d}, and object detection on Hypersim~\cite{roberts2021hypersim}, 3D-FRONT~\cite{fu20213d}, and ScanNet. The results show that classification and completion retain or improve accuracy, while for segmentation and detection, the method shows a favorable trade-off, exchanging modest accuracy for large computational savings. In all tasks, our method achieves substantial reductions in theoretical GFLOPs and improvements in measured on-device FPS compared to dense and sparse convolution baselines, demonstrating efficiency gains that enable real-time deployment. This confirms that our approach does not simply optimize accuracy, but instead maps the efficiency–accuracy trade-off across different tasks, which is critical for embedded robotic perception. On-device benchmarks on an NVIDIA Jetson Orin Nano further validate real-time throughput, supporting the practical relevance of the proposed design.
We summarize our main contributions as follows:
\begin{itemize}
    \item An interpolation-free tri-plane lifting pipeline that eliminates interpolation, point-wise queries, and implicit MLPs, establishing tri-planes as a practical backbone for 3D voxel reasoning.
    \item A hybrid architecture that integrates our lifting backbone with a low-resolution volumetric branch and an auxiliary positional encoding module that re-injects spatial context through data-adaptive modulation. This design effectively preserves global context and recovers fine details with negligible overhead while maintaining end-to-end GPU parallelism.
    \item A systematic evaluation and on-device validation across multiple 3D tasks, balancing computational costs with performance and confirming real-time throughput on an NVIDIA Jetson Orin Nano for embedded robotic perception.    
\end{itemize}

\section{Related work} 
Dense 3D convolution is highly expressive but its computational and memory costs scale cubically with voxel resolution. To mitigate this, submanifold and sparse convolutions along with MinkowskiNet confine computation exclusively to active, non-empty voxels, yielding substantial computational savings, particularly in environments with low occupancy~\cite{graham2017submanifold,graham20183d,choy20194d}. Later works continued to reduce redundancy and latency. SPS-Conv~\cite{liu2022spatial} prunes uninformative sites, VoxelNeXt~\cite{chen2023voxelnext} builds fully sparse backbones, and SparseOcc~\cite{tang2024sparseocc} maintains fully sparse latent volumes without loss of accuracy. Yet, these methods still use 3D kernels and add nontrivial overhead.

Orthogonal to sparsity, projection-based methods replace volumetric 3D convolution with 2D convolution after mapping 3D structure onto planar grids. Representative methods include OFT~\cite{roddick2018orthographic} for monocular lifting to an orthographic space, PointPillars~\cite{lang2019pointpillars}, which evolved from early sparse convolutional detectors like SECOND~\cite{yan2018second}, for pillarized pseudo images, RangeNet++\cite{milioto2019rangenet++} for range view segmentation, and Lift, Splat, Shoot\cite{philion2020lift} for multi-camera lifting to BEV, later extended by transformer-based methods such as BEVFormer~\cite{li2024bevformer}. More recently, EG3D~\cite{chan2022efficient} demonstrated that tri-plane features can support generative rendering, where features are interpolated from the planes and decoded by implicit MLPs. Triplane Diffusion~\cite{shue20233d} scaled this idea to diffusion-based 3D neural field generation, while SemCity~\cite{lee2024semcity} and TPVFormer~\cite{huang2023tri} adapted tri-planes for semantic scene synthesis and occupancy prediction, respectively. Although these methods show the versatility of tri-plane structures, they all rely on interpolation and point-wise queries coupled with implicit decoders, which introduce irregular memory access and limit real-time feasibility. In contrast, our design employs an interpolation-free broadcast-summation to directly reconstruct voxel features, enabling dense GPU-parallel execution and making it better suited for real-time 3D reasoning.

Projection-based pipelines collapse a 3D structure onto a set of 2D planes, a process that invariably discards critical spatial cues. Since tiling per-plane features back into 3D does not recover them, models inject positional information. Fixed encodings include sinusoidal mappings in Transformers~\cite{vaswani2017attention}, coordinate channels~\cite{liu2018intriguing}, and Fourier features~\cite{tancik2020fourier}, which are effective for high-frequency 3D representation learning~\cite{mildenhall2021nerf}. 

Learned variants extend this idea by adapting positional features from data, such as relative coordinate encodings in point clouds (Point Transformer~\cite{zhao2021point}) and discrete voxel embeddings (Voxel Transformer~\cite{mao2021voxel}). Multi-view detection frameworks further map 3D coordinates to embeddings with lightweight MLPs, as in PETR~\cite{liu2022petr}.
These approaches highlight the importance of positional cues for mitigating projection-induced losses, but they typically treat positional encoding as a central modeling component. In contrast, our method introduces an adaptive positional encoding as an auxiliary signal to strengthen interpolation-free tri-plane lifting.

\section{Method}
\begin{figure*}[t]
    \centering
    \includegraphics[width=0.98\textwidth]{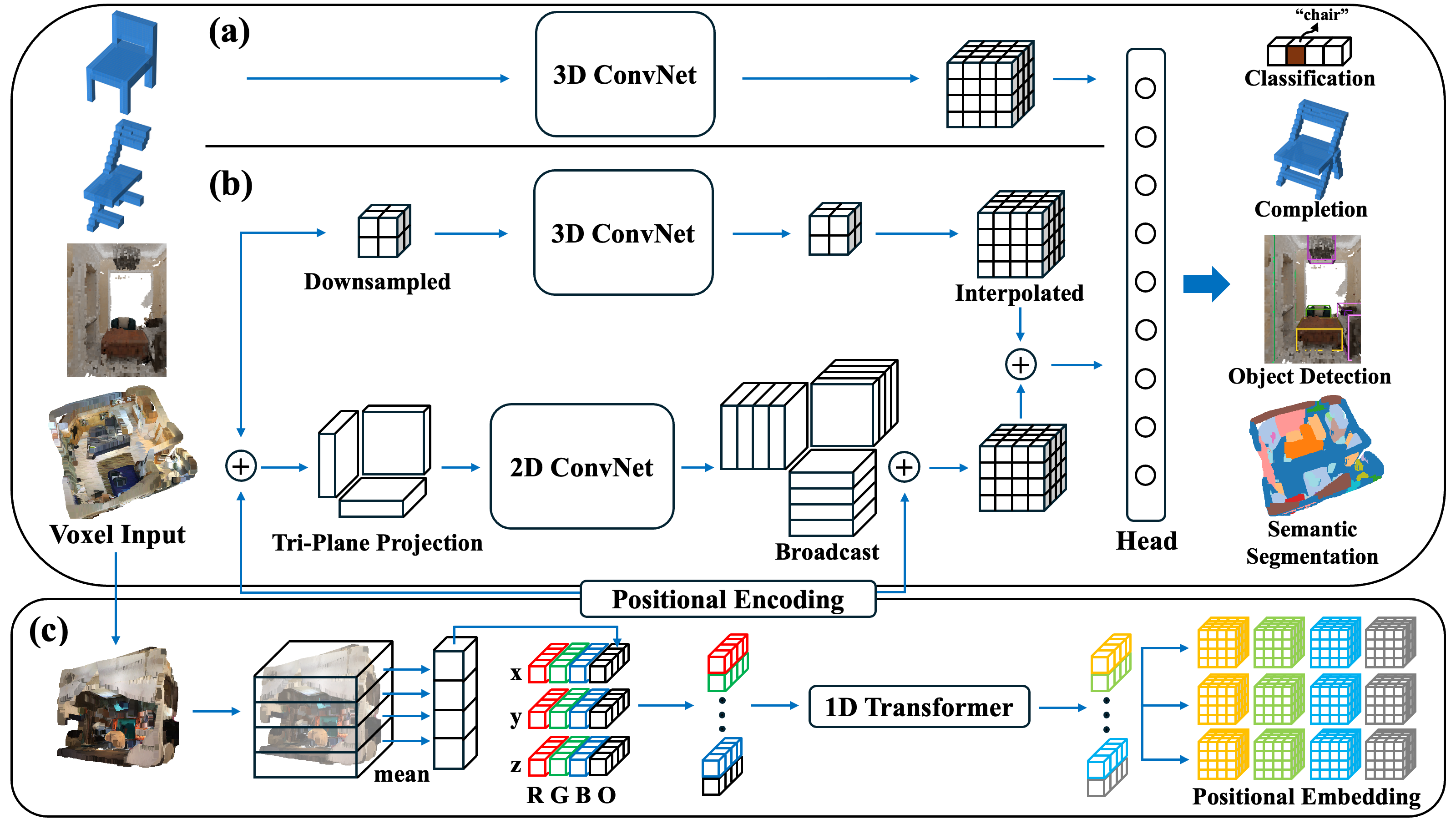}
    \caption{System Overview. (a) A dense 3D ConvNet baseline. (b) Our proposed hybrid architecture, which utilizes interpolation-free broadcast-summation to reconstruct feature volumes while shifting non-linear complexity to 2D convolutions for GPU acceleration. (c) The data-adaptive positional encoding process where 3D context is summarized into 1D tokens to generate dynamic spatial embeddings. These embeddings provide essential spatial cues to compensate for information loss and retain geometric details in our pipeline.
    }
    \label{fig:method}
    \vspace{-10pt}
\end{figure*}

We propose an efficient hybrid architecture to address the high computational cost of dense 3D convolutions. In the following subsections, we first detail our core interpolation-free tri-plane lifting framework, which includes an adaptive positional modulation to retain spatial detail (Section~\ref{subsec_me:A}). We then describe the parallel low-resolution volumetric branch used for capturing global context (Section~\ref{subsec_me:B}). Finally, we explain how these two streams are merged through a lightweight fusion module to produce the final representation (Section~\ref{subsec_me:C}).

\subsection{Interpolation-Free Tri-Plane Lifting}
\label{subsec_me:A}
The core of our approach is an interpolation-free lifting pipeline that constructs a 3D feature volume from 2D plane projections. This design shifts the main computational load to lightweight 2D convolutions, ensuring high efficiency and real-time performance. Figure~\ref{fig:method}a shows the dense 3D CNN baseline, whose full‑resolution convolutions incur heavy FLOPs and memory, while Figure~\ref{fig:method}b presents our proposed hybrid approach.

For the input volume $V \in \mathbb{R}^{C \times D_x \times D_y \times D_z}$, we create three orthogonal average projections, one for each primary axis $k\!\in\!\{x,y,z\}$, according to the following formula:
\begin{align}
    P_k(c,u,v)=\frac{1}{D_k}\sum_{d=1}^{D_k} V_k(c,u,v,d).
\end{align}
Each plane $P_k$ is processed by a lightweight 2D CNN $g_k$ to produce a feature map $F_k = g_k(P_k)$. We then lift these plane features back to 3D by broadcasting them along the missing dimension:
\begin{align}
\hat F_k(u,v,w)=F_k(u,v) \text{, where }k=x,y,z.
\end{align}
The lifted feature volume $T$ is then obtained as a weighted combination of the three tensors:
\begin{align}
    T=\lambda_x \hat{F}_x + \lambda_y \hat{F}_y + \lambda_z \hat{F}_z
\end{align}
where $\lambda_x, \lambda_y, \lambda_z$ are learnable coefficients that adaptively balance contributions from each plane.
Unlike prior tri-plane methods that rely on point-wise interpolation and implicit MLPs, our broadcast-summation scheme constructs the feature volume in one step, reducing the operation to dense element-wise additions that map efficiently to GPU kernels. This makes the process fully parallelizable, avoids irregular memory access, and scales linearly with voxel resolution rather than cubically as in dense 3D convolutions.

Projecting the 3D volume onto planes inevitably removes fine voxel-level relations, and simple broadcast-summation cannot fully recover them. To compensate, we introduce an adaptive positional modulation that supplements broadcast-addition with global spatial cues. A lightweight transformer summarizes the input volume $V \in \mathbb{R}^{C \times D_x \times D_y \times D_z}$ by average projections along the $x,y,z$ axes, producing token sequences $\mathbf{t}_{c,k}$. For example, the $x$-axis token is  
\begin{align}
\mathbf{t}_{c,x}[i] = \tfrac{1}{D_y D_z} \sum_{j=1}^{D_y} \sum_{l=1}^{D_z} V_{c, ijl}.
\end{align}
These tokens are processed to generate axis-conditioned embeddings $E=\{\mathbf{e}_{c,k}\}$, which act as weighting functions providing coarse global context.

The embeddings are applied as an additive bias in two stages. First, for pre-modulation, a weight volume $W_{pre}$ is generated from the embeddings and added to the input volume before projection:
\begin{align}
    V'_{c}(i,j,l) = V_{c}(i,j,l) + W_{pre}(i,j,l).
\end{align}
Accordingly, the projection then operates on the modulated volume $V'$ rather than the original $V$.
Second, for post-modulation, another weight volume $W_{post}$ is generated to refine the lifted volume $T$ after reconstruction, yielding the final volume $T' = T + W_{post}$. This two-stage modulation re-injects spatial cues with negligible overhead ($<$0.1\% FLOPs).

\subsection{Low-resolution Feature Volume}
\label{subsec_me:B}
In parallel, we downsample the input volume $V$ by a ratio $r$ to obtain $\tilde{V}$, and process $\tilde{V}$ with a compact 3D CNN $h$ to capture coarse global context. The result is then upsampled to the original resolution to form the volumetric feature $G$. This auxiliary pathway serves as a low-resolution volumetric branch. Following memory-efficient designs such as MeRF~\cite{reiser2023merf, duckworth2024smerf}, it applies 3D convolutions only on a reduced representation, minimizing overhead while retaining structural context.

Beyond efficiency, the low-resolution volumetric branch fulfills two key roles. First, it stabilizes training by injecting coarse structural cues that regularize the finer but noisier features of the lifted stream. Second, it reintroduces volumetric consistency that compensates for projection-induced information loss. Even at reduced resolution, this branch preserves long-range spatial correlations that are difficult to capture through tri-plane lifting alone, making it an essential complement to the broadcast-summation pathway.

\subsection{Module Fusion}
\label{subsec_me:C}
For fusing the lifted feature volume $T$ and volumetric features $G$, we opt for element-wise summation over concatenation, in order to avoid increasing channel depth. This simple fusion achieves comparable performance to concatenation while minimizing overhead.
While broadcast-summation alone can serve as the final fusion step, it limits representational flexibility.
To enrich the combined features without introducing heavy computation, motivated by \cite{he2016deep,redmon2016you}, we apply a 1\plh1\plh1 convolution block $\phi$, which provides an efficient nonlinear channel adjustment while preserving end-to-end GPU parallelism.
In our setting, it refines fused features without expanding spatial dimensions, keeping the operation lightweight while enhancing representational flexibility.
The output prediction $\hat{Y}$ is then obtained as:
\begin{align}
    \hat{Y} = \phi(T' + G).
\end{align}
Consequently, the two streams $T$ and $G$ are fused by element-wise summation and refined through a lightweight 1\plh1\plh1 convolutional block, producing a compact yet expressive representation.
The resulting volume is then passed to the downstream task-specific head for classification, completion, segmentation, or object detection.

\section{Experiments}
We evaluate TriLift across multiple 3D vision tasks to assess the efficiency–accuracy trade-offs of interpolation-free tri-plane lifting. Section~\ref{subsec_ex:A} describes the experimental setup, including datasets and evaluation metrics. Section~\ref{subsec_ex:B} reports comparative results against baseline methods on four 3D vision tasks. Section~\ref{subsec_ex:C} presents ablation studies that examine the role of each design component, including the auxiliary positional encoding module. Section~\ref{subsec_ex:D} measures on-device throughput on an NVIDIA Jetson Orin Nano to evaluate the method’s suitability for embedded deployment.

\subsection{Experimental Setup}
\label{subsec_ex:A}

\subsubsection{Datasets and Evaluation Metrics}
We evaluate our method on four 3D vision tasks: shape classification, shape completion, semantic segmentation, and object detection. For shape classification and completion, we use the ModelNet40~\cite{wu20153d} dataset. For semantic segmentation, we benchmark on two large-scale indoor scene datasets, ScanNet~\cite{dai2017scannet} and Stanford 3D~\cite{armeni20163d}. For our 3D object detection experiments, we use the same datasets as NeRF-RPN~\cite{hu2023nerf}: Hypersim~\cite{roberts2021hypersim}, 3D-FRONT~\cite{fu20213d}, and ScanNet~\cite{dai2017scannet}. For all tasks, we preprocess the raw data by converting it into a volumetric voxel grid representation. The grid resolution is set to $128^3$ for ModelNet40, ScanNet, and Stanford 3D. For the object detection datasets, we use a resolution of $160^3$ for 3D-FRONT and ScanNet, and $200^3$ for Hypersim to accommodate the varying scene scales. The input channel dimension $C$ is task-dependent: $C{=}1$ for binary occupancy in classification and completion, $C{=}1$ for NeRF density in object detection, and $C{=}4$ (occupancy with RGB) for semantic segmentation.

We employ task-specific metrics for performance evaluation. For classification, we report accuracy (Acc), F-score, and Area Under the Curve (AUC). For completion, performance is measured by L2 Chamfer distance, F-score, and Intersection over Union (IoU). In semantic segmentation, we use mean IoU (mIoU), accuracy, and F-score. Finally, for object detection, we report Recall (R) and Average Precision (AP) at IoU thresholds of 0.25 and 0.50.

\subsubsection{Implementation Details and Comparisons}
TriLift builds on an interpolation-free lifting backbone. We denote its minimal variant as \textit{TriLift-B} (backbone), which projects the input volume onto three orthogonal planes, applies lightweight $2D$ convolutions, and restores the $3D$ feature volume via broadcast-summation.
To assess the impact of auxiliary components, we further report \textit{TriLift-F} (Full), which augments the backbone with adaptive positional encoding \textit{(PE)}, a low-resolution volumetric branch \textit{(Vol)}, and a 1\plh1\plh1 channel mixer. We also evaluate variants, denoted as \textit{TriLift-F(r)}, where the volumetric branch processes an input downsampled by a ratio of r, to illustrate the efficiency–accuracy trade-off. For ablations, we define \textit{TriLift-F(NoPE)} as the full model without \textit{PE}, and \textit{TriLift-F(var)} as versions where \textit{PE} is replaced by sinusoidal, coordinate, or MLP encodings. These comparisons allow us to quantify the role of each component and to demonstrate that the full variant achieves the best trade-off among accuracy, efficiency, and real-time feasibility.

We compare our proposed architecture with other representative baselines. The first is a \textit{3D Base} model, which represents a standard dense 3D convolutional neural network using a U-Net-like architecture. It processes the entire voxel grid and serves as a benchmark for maximum expressive power at a high computational cost. The second baseline is \textit{3D Sparse}, which is designed to exploit the inherent spatial sparsity within 3D data by building on submanifold sparse convolutions, similar to methods like MinkowskiNet~\cite{choy20194d}. It restricts computations to active voxels to reduce FLops.

For the segmentation task, both streams utilize U-Net architectures to effectively handle dense prediction. For the other tasks of classification, completion, and object detection, we use standard feed-forward CNNs.
All models were trained on a single NVIDIA RTX 4090 GPU.

\begin{table}[t]
  \centering
  \begin{threeparttable}
  \caption{3D Classification Performance Comparison on ModelNet40}
  \label{tab:cls_zero}
  \setlength{\tabcolsep}{11pt}
  \begin{tabular}{@{}>{\hspace{1em}}l|cccc@{\hspace{1em}}}
    \toprule
    Method       & GFLOPs & Acc   & F-score & AUC     \\ 
    \midrule
    3D Base      & 29.38  & \cellcolor{orange!40}83.94   & 75.74   & \cellcolor{orange!20}0.9871  \\
    3D Sparse    & 6.34   & \cellcolor{orange!20}83.54 & 74.92   & \cellcolor{orange!40}0.9877  \\
    \textit{TriLift-B}  & \cellcolor{orange!40}0.96   & 82.52 & \cellcolor{orange!20}75.86 & 0.9857  \\
    \textit{TriLift-F($1/2$)}   & 4.42   & 82.93 & 75.81   & 0.9844  \\
    \textit{TriLift-F($1/4$)}   & \cellcolor{orange!20}1.34 & \cellcolor{orange!20}83.54 & \cellcolor{orange!40}76.60   & 0.9823  \\
    \bottomrule
  \end{tabular}
  The \colorbox{orange!40}{best} and \colorbox{orange!20}{second-best} results are highlighted.
  \end{threeparttable}
\end{table}


\begin{table}[t]
  \centering
  \begin{threeparttable}
  \caption{3D Completion Performance Comparison on ModelNet40}
  \label{tab:completion}
  \setlength{\tabcolsep}{9pt}
  \begin{tabular}{@{}>{\hspace{1em}}l|cccc@{\hspace{1em}}}
    \toprule
    Method            & GFLOPs & L2 Chamfer      & F-score & IoU     \\ 
    \midrule
    3D Base           & 78.61  & \cellcolor{orange!20}0.001421  & 70.87   & 56.22   \\ 
    3D Sparse         & 29.52  & 0.002205        & 66.27   & 50.20   \\ 
    \textit{TriLift-B}       & \cellcolor{orange!40}2.05   & 0.012340        & 49.26   & 33.08   \\ 
    \textit{TriLift-F($1/2$)}        & 12.28  & 0.001494        & \cellcolor{orange!40}76.26   & \cellcolor{orange!40}62.03   \\ 
    \textit{TriLift-F($1/4$)}        & \cellcolor{orange!20}3.59   & \cellcolor{orange!40}0.001392  & \cellcolor{orange!20}74.23   & \cellcolor{orange!20}60.07   \\ 
    \bottomrule
  \end{tabular}
  \end{threeparttable}
\end{table}

\subsection{Experiment Results and Analysis}
\label{subsec_ex:B}
\begin{table}[t]
  \centering
  \begin{threeparttable}
  \caption{3D Semantic Segmentation on ScanNet and Stanford 3D}
  \label{tab:semseg_three_zero}
  \setlength{\tabcolsep}{4pt}
  \begin{tabular}{@{}l|c|ccc|ccc@{}}
    \specialrule{\heavyrulewidth}{0pt}{1pt}
    \multirow{2}{*}{Method}
      & \multirow{2}{*}{GFLOPs}
      & \multicolumn{3}{c|}{ScanNet}
      & \multicolumn{3}{c}{Stanford 3D} \\
      & 
      & mIoU  & Acc   & F-score
      & mIoU  & Acc   & F-score \\
    \midrule
    3D Base     & 540.18 & \cellcolor{orange!20}40.21 & \cellcolor{orange!20}74.68 & \cellcolor{orange!20}60.34
                & \cellcolor{orange!40}51.63 & \cellcolor{orange!20}84.14 & \cellcolor{orange!40}68.47 \\
    3D Sparse   & 118.70 & \cellcolor{orange!40}42.17 & \cellcolor{orange!40}79.34 & \cellcolor{orange!40}61.98
                & 48.90 & \cellcolor{orange!40}85.38 & 66.25 \\
    \textit{TriLift-B} & \cellcolor{orange!40}10.35 & 24.12 & 55.50 & 43.01
                & 40.46 & 73.44 & 60.91 \\
    \textit{TriLift-F($1/2$)}  & 87.21 & 34.19 & 70.50 & 54.62
                & \cellcolor{orange!20}50.53 & 82.18 & \cellcolor{orange!20}68.28 \\
    \textit{TriLift-F($1/4$)}  & \cellcolor{orange!20}27.63 & 30.18 & 63.78 & 49.44
                & 44.69 & 76.36 & 61.59 \\
    \specialrule{\heavyrulewidth}{1pt}{0pt}
  \end{tabular}
  \end{threeparttable}
\end{table}

\begin{figure*}[t]
    \centering
    \includegraphics[width=1.00\textwidth]{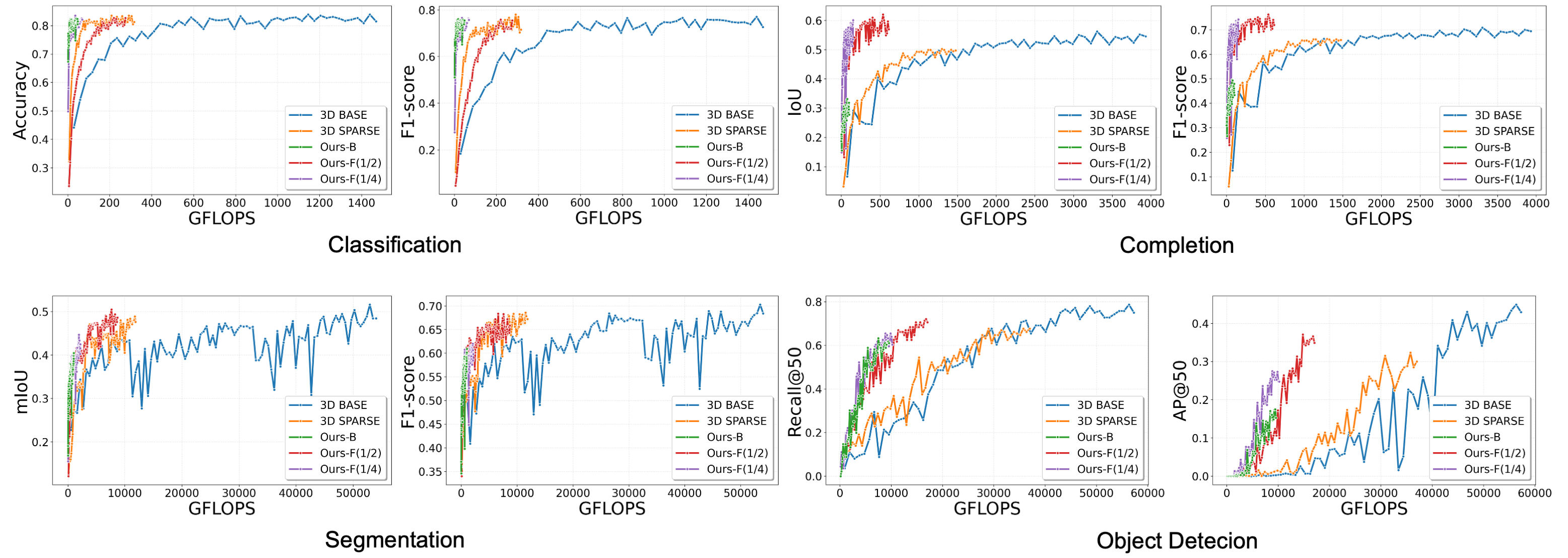}
    \caption{Performance against computational cost across four 3D vision tasks. Our full models, \textit{TriLift-F}(1/2) and \textit{TriLift-F}(1/4), achieve accuracy comparable to the computationally intensive \textit{3D Base} and \textit{3D Sparse} models at a fraction of the computational cost. The \textit{TriLift-B} variant highlights the efficiency of our core backbone, which provides a strong baseline. Notably, for the shape completion task, our approach surpasses the performance of all baseline methods.
    }
    \label{fig:graph}
\end{figure*}

Overall, our hybrid architecture achieves a balanced trade-off between computational efficiency and task performance. As illustrated in Figure~\ref{fig:graph}, the models \textit{TriLift-F}(1/2) and \textit{TriLift-F}(1/4) occupy favorable positions along the efficiency–accuracy spectrum. They retain or improve accuracy in tasks such as classification and completion while substantially reducing GFLOPs, and they show a trade-off with moderate performance drops in exchange for large computational savings. This validates our design: interpolation-free lifting secures efficiency, while positional encoding serves as an auxiliary bridge that helps mitigate, though not completely eliminate, the information loss from projection.

The classification results in Table~\ref{tab:cls_zero} show that performance differences between methods are minor, as the task does not strictly require complex 3D reasoning. This performance gap widens significantly in shape completion, detailed in Table~\ref{tab:completion}, where full 3D understanding is crucial. \textit{TriLift-F}(1/4) surpasses the heavyweight \textit{3D Base} model across all metrics, achieving an F-score of 74.23 and an IoU of 60.07, compared to the baseline's scores of 70.87 and 56.22, respectively.
We interpret this outcome not as a universal accuracy gain, but as an effect of the hybrid architecture acting as an information bottleneck, which provides a regularization-like benefit in this specific task.
This process is enabled by our data-adaptive positional encoding, which bridges the information lost during projection while keeping the representation compact, leading to favorable trade-offs in tasks that require strong geometric reasoning.

In semantic segmentation, the results in Table~\ref{tab:semseg_three_zero} highlight the efficiency-accuracy trade-offs of our approach. On the Stanford 3D dataset, \textit{TriLift-F}(1/2) model achieves an mIoU of 50.53, outperforming the 3D Sparse baseline (48.90) while operating at lower computational cost, and approaching the performance of the dense 3D Base model (51.63). These results suggest that even though projection-based lifting inevitably loses some voxel-level relationships, our data-adaptive positional encoding partially compensates for this loss and yields competitive accuracy when the dataset structure is favorable. At the same time, the efficiency advantage of our method becomes clear in object detection, where dense inputs reduce the benefit of sparse convolution. As shown in Table~\ref{tab:rpn_three_zero}, the 3D Sparse model requires 1262.09 GFLOPs on Hypersim, close to the 1615.84 GFLOPs of the 3D Base, while \textit{TriLift-F}(1/4) operates at only 77.65 GFLOPs. Despite this drastic reduction, it delivers detection performance competitive with both baselines, confirming that our design consistently provides an effective trade-off between efficiency and accuracy.


\begin{table*}[t]
  \centering
  \begin{threeparttable}
  \caption{3D Object Detection Performance Comparison across 3D-FRONT, Hypersim, and ScanNet}
  \label{tab:rpn_three_zero}
  \setlength{\tabcolsep}{4pt}
  \begin{tabular}{@{}>{\hspace{1em}}l|ccccc|ccccc|ccccc@{\hspace{1em}}}
    \specialrule{\heavyrulewidth}{0pt}{1pt}
    \multirow{2}{*}{Method}
      & \multicolumn{5}{c|}{3D-FRONT}
      & \multicolumn{5}{c|}{Hypersim}
      & \multicolumn{5}{c}{ScanNet} \\
      
      & FLOPs & R@25 & R@50 & A@25 & A@50
      & FLOPs & R@25 & R@50 & A@25 & A@50
      & FLOPs & R@25 & R@50 & A@25 & A@50 \\
    \midrule
    3D Base       & 823.44 & \cellcolor{orange!40}98.53 & \cellcolor{orange!40}78.68 & \cellcolor{orange!40}65.89 & \cellcolor{orange!40}44.93
                  & 1615.84 & \cellcolor{orange!40}65.40 & \cellcolor{orange!40}25.71 & \cellcolor{orange!40}14.16 & \cellcolor{orange!40}4.04
                  & 823.44 & \cellcolor{orange!40}89.16 & \cellcolor{orange!40}42.36 & \cellcolor{orange!40}38.47 & \cellcolor{orange!40}10.51 \\
    3D Sparse     & 485.16 & \cellcolor{orange!20}97.79 & 67.65 & 53.57 & 24.90
                  & 1262.09 & 54.33 & 13.33 & 5.33 & 0.68
                  & 395.44 & 82.37 & 26.69 & 16.87 & 1.55 \\
    \textit{TriLift-B}   & \cellcolor{orange!40}28.21 & 94.85 & 63.24 & 31.86 & 11.91
                  & \cellcolor{orange!40}44.44 & 52.02 & 11.38 & 3.29 & 0.48
                  & \cellcolor{orange!40}28.21 & 81.77 & 22.17 & 19.36 & 1.35 \\
    \textit{TriLift-F($1/2$)}    & 152.95 & 97.06 & \cellcolor{orange!20}72.06 & \cellcolor{orange!20}61.92 & \cellcolor{orange!20}36.62
                  & 305.90 & \cellcolor{orange!20}60.32 & \cellcolor{orange!20}15.68 & \cellcolor{orange!20}6.93 & \cellcolor{orange!20}1.07
                  & 152.95 & \cellcolor{orange!20}88.67 & \cellcolor{orange!20}36.45 & \cellcolor{orange!20}21.08 & \cellcolor{orange!20}4.26 \\
    \textit{TriLift-F($1/4$)}    & \cellcolor{orange!20}38.82 & 95.59 & 65.44 & 55.20 & 27.46
                  & \cellcolor{orange!20}77.65 & 53.06 & 13.97 & 5.82 & 0.58
                  & \cellcolor{orange!20}38.82 & 85.71 & 30.05 & 18.55 & 2.01 \\
    \specialrule{\heavyrulewidth}{1pt}{0pt}
  \end{tabular}
    FLOPs are reported in GFLOPs. R@25 and R@50 denote Recall, while A@25 and A@50 denote Average Precision, evaluated at IoU thresholds of 0.25 and 0.50, respectively. The reported FLOPs measure the 3D backbone and FPN. All methods share a common 3D RPN head, which adds a fixed computational cost of 131.58 GFLOPs for 3D-FRONT, 257.61 GFLOPs for Hypersim, and 131.58 GFLOPs for ScanNet.
  \end{threeparttable}
\end{table*}

\begin{figure*}[t]
    \centering
    \includegraphics[width=0.99\textwidth]{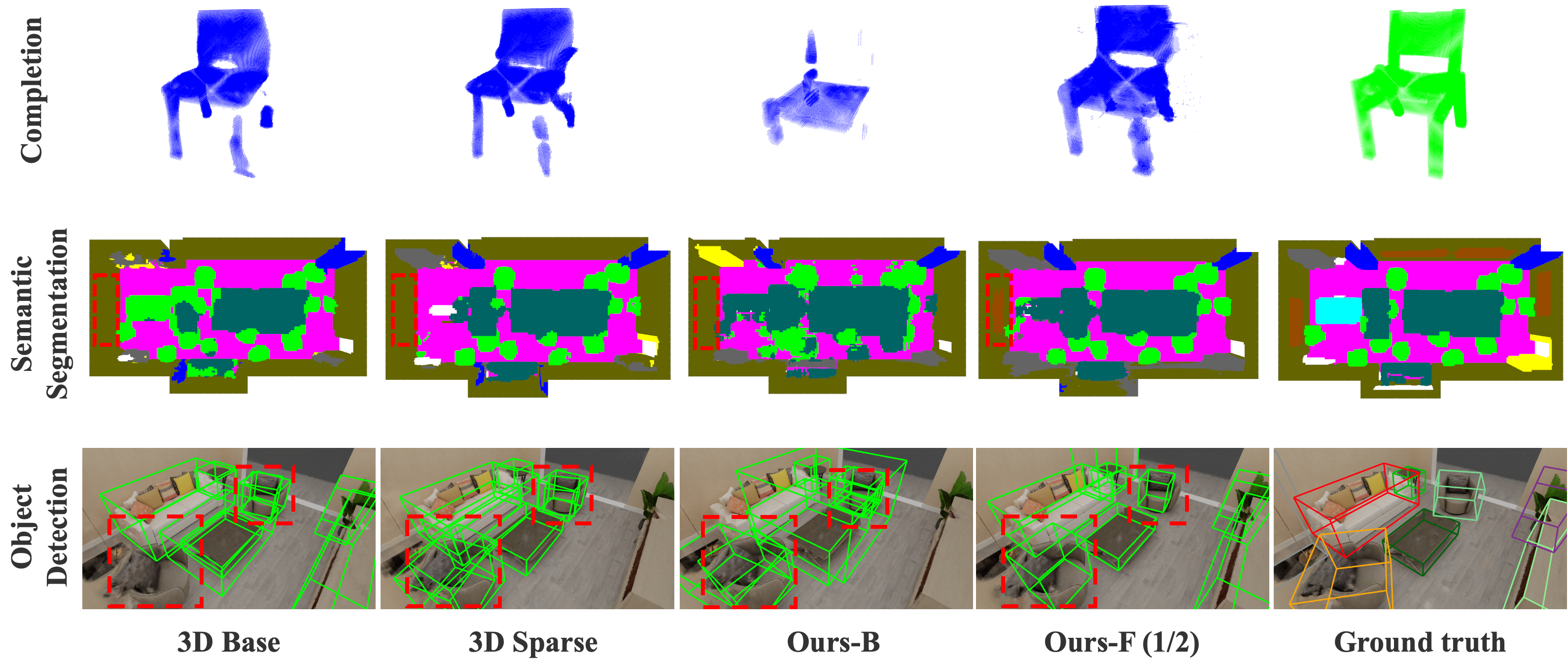}
    \caption{Qualitative results for shape completion, semantic segmentation, and object detection. The visualization shows results for completion on ModelNet40~\cite{wu20153d}, semantic segmentation on Stanford3D~\cite{armeni20163d}, and object detection on 3D-FRONT~\cite{fu20213d}. Our full model, labeled \textit{TriLift-F(1/2)}, reconstructs a more complete shape in completion, while producing cleaner segmentation maps and more accurate bounding boxes than our backbone model, \textit{TriLift-B}. The visual quality is comparable to the more computationally expensive 3D methods.}
    \label{fig:qualitative}
\end{figure*}

We provide a qualitative analysis in Figure~\ref{fig:qualitative} to assess our method's performance. For shape completion, the limitations of projection-based methods are evident; the \textit{TriLift-B} baseline produces a sparse, incomplete artifact that fails to capture the object's structure. While the \textit{3D Base} and \textit{3D Sparse} models generate more reasonable shapes, they struggle with fine details, leaving chair legs partially missing or distorted. In contrast, our method reconstructs a complete chair with all four legs, closely matching the ground truth geometry.
This consistency is observed in semantic segmentation. The \textit{TriLift-B} output is heavily fragmented, and even the \textit{3D Base} and \textit{3D Sparse} methods show noisy boundaries. As highlighted in the red dashed box on the left, other methods mis-segment the object, whereas our approach produces a cleaner map that aligns more closely with the ground truth.
A similar pattern appears in object detection, where the other methods either miss objects or produce multiple inaccurate boxes for a single object, as highlighted by the red dashed boxes. Our method instead produces bounding boxes that are more consistent and better aligned with the ground truth. These visualizations indicate that our architecture, despite its lower computational cost, retains structural fidelity that is often degraded in other approaches.


\subsection{Positional Encoding Variants}
\label{subsec_ex:C}
Positional encoding introduces consistent accuracy gains across tasks at virtually no additional cost, with overheads below 0.1\% of total FLOPs. These moderate improvements nonetheless represent an economical enhancement that reliably strengthens reconstructed features without compromising efficiency. Table~\ref{tab:ablation_zero} compares the full model (\textit{TriLift-F}(1/2)) with several variants: one without positional encoding, and others using fixed sinusoidal encoding~\cite{vaswani2017attention}, coordinate channels (CoordConv)\cite{liu2018intriguing}, or a learned MLP\cite{liu2022petr}. Our transformer-based positional encoding module consists of a 2-layer Transformer Encoder with 8 attention heads and a model dimension of 128. The model without any positional encoding yields the lowest scores across nearly all metrics, such as an F1 of 74.38 in completion, indicating that bridging spatial information is important for counteracting the loss from projection and reconstruction. Introducing fixed encodings like \textit{Sinusoidal} and \textit{CoordConv} gives only marginal benefit, reflecting the limited expressiveness of static mappings. The learned \textit{MLP} variant generally performs better, for instance reaching an F1 of 67.63 on Stanford3D, suggesting that learned embeddings provide more flexibility. The transformer-based encoding achieves the strongest results overall, such as 54.62 F1 on ScanNet segmentation compared to 53.52 for the MLP-based variant. These results show that context-aware positional signals generated by the transformer, which capture global structure and inter-slice relationships, consistently yield the most reliable improvements across tasks.

\begin{table*}[t]
  \centering
  \begin{threeparttable}
  \caption{Ablation study of positional encodings and architectures across tasks and datasets}
  \label{tab:ablation_zero}
  \setlength{\tabcolsep}{10pt}
  \begin{tabular}{@{}l|cccccccc@{}}
    \toprule
    \multirow{2}{*}{Method} &
    \multicolumn{1}{c}{Classification} &
    \multicolumn{1}{c}{Completion} &
    \multicolumn{2}{c}{Semantic Segmentation} &
    \multicolumn{3}{c}{Object Detection} \\
    \cmidrule(lr){2-2}\cmidrule(lr){3-3}\cmidrule(lr){4-5}\cmidrule(lr){6-8}
     & ModelNet & ModelNet & ScanNet & Stanford3D & Hypersim & 3D-FRONT & ScanNet \\
    \midrule
    \textit{TriLift-F(NoPE)} & 74.55 & 74.38 & 53.70 & 65.41 & 70.12 & \cellcolor{orange!20}14.60 & 33.53 \\
    \textit{TriLift-F(Sinusoidal)}             & 74.31 & \cellcolor{orange!20}76.04 & \cellcolor{orange!20}54.08 & 66.65 & 68.44 & 13.43 & 34.54 \\
    \textit{TriLift-F(CoordConv)}              & 74.42 & 74.39 & 53.38 & 66.11 & 68.38 & 13.73 & 34.47 \\
    \textit{TriLift-F(MLP)}                    & \cellcolor{orange!20}74.97 & 75.64 & 53.52 & \cellcolor{orange!20}67.63 & \cellcolor{orange!20}70.59 & 14.33 & \cellcolor{orange!40}36.51 \\
    \textit{TriLift-F(Transformer)}            & \cellcolor{orange!40}75.81 & \cellcolor{orange!40}76.26 & \cellcolor{orange!40}54.62 & \cellcolor{orange!40}68.28 & \cellcolor{orange!40}72.06 & \cellcolor{orange!40}15.68 & \cellcolor{orange!20}36.45 \\
    \bottomrule
  \end{tabular}
  Performance is evaluated using the F1-score for classification, completion, and segmentation, and Recall@50 for object detection. All results are reported for the \textit{TriLift-F}(1/2) model configuration.
  \end{threeparttable}
\end{table*}


\subsection{On-Device Performance Evaluation}
\label{subsec_ex:D}
To validate the practical applicability of our method for real-world robotics, we extend our analysis beyond theoretical FLOPs and measure the on-device inference speed. We deployed our models on the onboard NVIDIA Jetson Orin Nano with a RealSense D435i depth camera. The results, reported as Frames Per Second (FPS) for each task, are summarized in Table~\ref{tab:fps_measurement}.

The on-device results confirm the efficiency of our hybrid approach, consistently outperforming the dense 3D Base model across all tasks. Moreover, unlike sparse convolution baselines that faced deployment issues on ARM-based hardware due to specialized kernel dependencies, our framework ensures seamless portability by relying on standard tensor operations. 
We further verified real-world robustness in our on-device setting using a handheld RealSense camera. Despite the noise and irregular motion of handheld operation, our framework consistently produced stable 3D perception. 

Although performance metrics cannot be computed for these real-world sequences due to the absence of ground truth, the model architecture and weights remain identical to the workstation setup, so the accuracy and efficiency–accuracy balance reported in Tables \ref{tab:cls_zero}, \ref{tab:completion}, \ref{tab:semseg_three_zero}, and \ref{tab:rpn_three_zero} are preserved on the Jetson Orin Nano.

However, we observe that the measured FPS gains, while substantial, are not as pronounced as the theoretical GFLOPs reductions would suggest. This discrepancy arises primarily because the system is limited by memory on hardware with restricted resources, where the latency of 3D voxel data access becomes the dominant bottleneck. Furthermore, our current implementation processes the tri-plane and volumetric streams sequentially, presenting a clear opportunity for future optimization through parallel processing to better leverage multi-core embedded capabilities.

\begin{table}[t]
  \centering
  \begin{threeparttable}
  \caption{On-device FPS for Each Task on NVIDIA Jetson Orin Nano}
  \label{tab:fps_measurement}
  \setlength{\tabcolsep}{12pt} 
  \begin{tabular}{l|cccc}
    \specialrule{\heavyrulewidth}{0pt}{1pt}
    Method & Cls. & Comp. & Seg. & OD \\
    \midrule
    3D Base     & 14.20 & 11.28 & 1.53 & 1.85 \\
    \textit{TriLift-B} & \cellcolor{orange!40}49.58 & \cellcolor{orange!40}35.79 & \cellcolor{orange!40}10.87 & \cellcolor{orange!40}10.64 \\
    \textit{TriLift-F}($1/2$)  & 24.75 & 14.86 & 3.09 & 3.16 \\
    \textit{TriLift-F}($1/4$)  & \cellcolor{orange!20}45.93 & \cellcolor{orange!20}17.52 & \cellcolor{orange!20}4.01 & \cellcolor{orange!20}6.09 \\    
    \specialrule{\heavyrulewidth}{1pt}{0pt}
  \end{tabular}
  Cls., Comp., Seg., and OD stand for Classification, Completion, Segmentation, and Object Detection, respectively. 
  \end{threeparttable}
\end{table}


\section{CONCLUSION}
In this work, we addressed the fundamental challenge of balancing computational cost and accuracy in 3D perception. We introduced TriLift, an efficient hybrid architecture that integrates a full-resolution 2D tri-plane stream with a low-resolution 3D volumetric stream, fused through a lightweight integration layer for end-to-end GPU acceleration. Specifically, an adaptive positional encoding module is incorporated to effectively mitigate spatial information loss, dynamically recovering geometric details with negligible overhead. Our extensive experiments across classification, completion, segmentation, and object detection show that the method achieves substantial computational savings while mapping the performance trade-off across tasks: accuracy is maintained or improved for classification and completion, while for segmentation and detection, the method achieves significant efficiency gains with only a slight decrease in accuracy. These results demonstrate the practical relevance of our design by enabling real-time throughput on embedded hardware, supporting its applicability to robotic perception under resource constraints.

\bibliographystyle{IEEEtran}
\bibliography{main}

\end{document}